\pdfoutput=1

\documentclass[11pt]{article}

\usepackage[]{EAMT23}

\usepackage{times}
\usepackage{url}
\usepackage{latexsym}

\usepackage[T1]{fontenc}

\usepackage[utf8]{inputenc}

\usepackage{microtype}

\usepackage{inconsolata}

\usepackage{xcolor}
\usepackage{lipsum}
\usepackage{booktabs}
\usepackage{multicol}
\usepackage{multirow}
\usepackage{graphicx}
\usepackage{tabularx}
\usepackage{bm}
\usepackage{float}
\usepackage{todonotes}
\usepackage{xargs}
\usepackage{xspace}
\usepackage{amsmath}
\usepackage{cleveref}
\usepackage{subcaption}
\usepackage{mathtools}
\usepackage{bigstrut}
\usepackage{tikz}
\usetikzlibrary{shapes}
\usepackage{rotating}
\usepackage{siunitx}
\usepackage{amssymb}
\usepackage{arydshln}
\usepackage{listings}
\usepackage[inline]{enumitem}
\usepackage{acronym}
\usepackage[parfill]{parskip} 
\usepackage{caption}
\usepackage{scrextend}
\usepackage{balance}
\usepackage{nccmath}

\newcommand*{\eg}{e.g.\@\xspace}
\newcommand*{\ie}{i.e.\@\xspace}

\makeatletter\newcommand*{\etc}{%
	\@ifnextchar{.}%
	{etc}%
	{etc.\@\xspace}%
}
\makeatother

\newcommand{\lde}{\textsc{De}}
\newcommand{\len}{\textsc{En}}
\newcommand{\lro}{\textsc{Ro}}

\newcommand{\adam}{\textsc{Adam}\@\xspace}

\newcommand{\lende}{\len-\lde}
\newcommand{\lenro}{\len-\lro}
\newcommand{\ldeen}{\lde-\len}
\newcommand{\lroen}{\lro-\len}

\newcommand{\sfour}{\textsc{S4}}
\newcommand{\sfourbi}{\textsc{S4bi}}
\newcommand{\sfoura}{\textsc{S4a}}
\newcommand{\tr}{\textsc{Tr}}

\definecolor{red}{RGB}{141,45,57}
\definecolor{dark}{RGB}{55,65,74}
\definecolor{blue}{RGB}{0,105,170}
\definecolor{gold}{RGB}{174,159,109}
\definecolor{gray}{RGB}{175,179,183}
\definecolor{darkgreen}{RGB}{50,110,30}

\definecolor{ptgreen}{RGB}{213,232,212}

\makeatletter
\newcommand*\standardbin{+}
\newcommand*\tabularbin[1]{%
  \mathbin{\mathpalette{\@tabularsym\standardbin}{#1}}%
}
\newcommand*\@tabularsym[3]{%
  \setbox\z@\hbox{$#2#1\m@th$}%
  \hbox to\wd\z@{\hss$#2#3\m@th$\hss}%
}
\makeatother

\newcolumntype{Y}{>{\centering\arraybackslash}X}

\newcommand{\bleu}{\textsc{Bleu}\xspace}

\newcommand{\sacrebleu}{\texttt{sacreBLEU}\xspace}

\newcommandx{\robin}[2][1=]{\vspace{0.2cm} \todo[linecolor=blue,backgroundcolor=blue!15,bordercolor=blue, #1]{\textbf{Robin:} #2}}
\newcommandx{\telmo}[2][1=]{\vspace{0.2cm} \todo[linecolor=gold,backgroundcolor=gold!25,bordercolor=gold, #1]{\textbf{Telmo:} #2}}
\newcommandx{\ali}[2][1=]{\vspace{0.2cm} \todo[linecolor=darkgreen,backgroundcolor=darkgreen!25,bordercolor=darkgreen, #1]{\textbf{Ali:} #2}}

\newcommand{\todotemplate}{\textcolor{red}{TODO}\xspace}
\newcommandx{\trobin}[2][1=]{\vspace{0.2cm} \todo[linecolor=blue,backgroundcolor=blue!15,bordercolor=blue, #1]{\textbf{\todotemplate @Robin:} #2}}
\newcommandx{\ttelmo}[2][1=]{\vspace{0.2cm} \todo[linecolor=gold,backgroundcolor=gold!25,bordercolor=gold, #1]{\textbf{\todotemplate @Telmo:} #2}}
\newcommandx{\tstephan}[2][1=]{\vspace{0.2cm} \todo[linecolor=darkgreen,backgroundcolor=darkgreen!25,bordercolor=darkgreen, #1]{\textbf{\todotemplate @Ali:} #2}}

\makeatletter
\def\adl@drawiv#1#2#3{%
        \hskip.5\tabcolsep
        \xleaders#3{#2.5\@tempdimb #1{1}#2.5\@tempdimb}%
                #2\z@ plus1fil minus1fil\relax
        \hskip.5\tabcolsep}
\newcommand{\cdashlinelr}[1]{%
  \noalign{\vskip\aboverulesep
           \global\let\@dashdrawstore\adl@draw
           \global\let\adl@draw\adl@drawiv}
  \cdashline{#1}
  \noalign{\global\let\adl@draw\@dashdrawstore
           \vskip\belowrulesep}}
\makeatother

\acrodef{MT}{machine translation}
\acrodef{LM}{language modeling}
\acrodef{S4}{Structured State Spaces for Sequences}

\title{State Spaces Aren't Enough: Machine Translation Needs Attention}

\author{\hspace{-2.5cm}Ali Vardasbi$^{\dagger *}$ \\ \hspace{-2.5cm} University of Amsterdam \\ \hspace{-2.5cm} \texttt{a.vardasbi@uva.nl}
\And  \hspace{-2cm} Telmo Pessoa Pires$^\dagger$ \; Robin M.~Schmidt \; Stephan Peitz \\ \hspace{-2cm} Apple \\
\hspace{-2cm} \texttt{\{telmo, robin\_schmidt, speitz\}@apple.com}
}

\begin{document}
\maketitle
\def\thefootnote{$\dagger$}\footnotetext{Equal contribution.}\def\thefootnote{\arabic{footnote}}
\def\thefootnote{$*$}\footnotetext{Work done during an internship at Apple.}\def\thefootnote{\arabic{footnote}}

\begin{abstract}
Structured State Spaces for Sequences (S4) is a recently proposed sequence model with successful applications in various tasks, e.g. vision, language modeling, and audio. Thanks to its mathematical formulation, it compresses its input to a single hidden state, and is able to capture long range dependencies while avoiding the need for an attention mechanism. In this work, we apply S4 to Machine Translation (MT), and evaluate several encoder-decoder variants on WMT'14 and WMT'16.
In contrast with the success in language modeling, we find that S4 lags behind the Transformer by approximately $4$ \bleu points, and that it counter-intuitively struggles with long sentences. Finally, we show that this gap is caused by S4's inability to summarize the full source sentence in a single hidden state, and show that we can close the gap by introducing an attention mechanism.

\end{abstract}

\section{Introduction}
\label{sec:intro}
The Transformer \cite{VaswaniSPUJGKP17} is the most popular architecture for state-of-the-art Natural Language Processing (NLP) \cite{devlin-etal-2019-bert,BrownMRSKDNSSAA20,NLLB}.
However, the attention mechanism on which it is built is not well suited for capturing long-range dependencies due to its quadratic complexity  \cite{ma2023mega}.
Recently, Structured State Spaces for Sequences (S4) was shown to be on par with the Transformer on various sequence modelling tasks, including time series forecasting, language modeling \cite{gu2022efficiently}, and audio generation \cite{GoelGDR22}; and to surpass the Transformer on tasks requiring reasoning over long range dependencies, like the \emph{Long Range Arena} \cite{Tay0ASBPRYRM21}.

Internally, S4 keeps a state-space based representation. Due to the way its weights are initialized, it is able to approximately ``memorize'' the input sequence, removing the need for an attention mechanism. Indeed, the results from \newcite{gu2022efficiently} show that the self-attention layers can be replaced by S4 layers without losing accuracy, and that it is able to effectively model long-range dependencies in data.
Moreover, one of the key advantages of the S4 kernel is that its forward step can be formulated both as a convolution and as a recurrence formula, allowing fast implementation during training, when the convolution method is used, while the recurrence formula is used to generate the output step by step during inference.

S4’s competitive performance in Language Modeling (LM) promises an alternative to the Transformer for other sequence modeling tasks, such as Machine Translation (MT).
In this work, we explore S4-based architectures for MT. Our goal is to find the best performing S4 architecture, and we study the impact of several architectural choices on translation accuracy, namely the effect of model depth, the number of S4 blocks, and the importance of the encoder.
Despite our best efforts, our top performing attention-free S4 model lags significantly ($\sim 4$ \bleu points) behind the Transformer, with the gap increasing with input length. We hypothesize this is due to the fact that S4 compresses the source sentence to a fixed-size representation, and thus lacks a way to access the token-level states of the source, which is important for MT.
As the input length increases, it becomes increasingly hard for the model to accurately store the full source sentence in a single hidden state. In contrast, the decoder cross-attention in the Transformer acts as a retrieval mechanism, allowing to accurate retrieval of the source sentence during decoding. Armed with this observation, we enhance S4 with cross-attention, and show this is enough to close the gap to the Transformer. Finally, we combine the Transformer and S4 into an hybrid architecture that outperforms both of them.

To summarize, the main contributions of the present work are:
\begin{enumerate}
    \item We present an in-depth study of S4 for MT.
    \item We provide evidence that S4 learns \emph{self-dependencies}, i.e. dependencies between the tokens of a single sequence, but struggles to capture \emph{cross-dependencies}, i.e. dependencies between the tokens of two sequences, as it lacks a way to retrieve prior states.
    \item We show that extending S4 with an attention mechanism allows it to more accurately capture cross-dependencies and to close the gap to the Transformer on MT.
\end{enumerate}

\begin{figure*}[t]
    \centering
    \begin{subfigure}{0.49\textwidth}
    \centering
    \includegraphics[width=\textwidth]{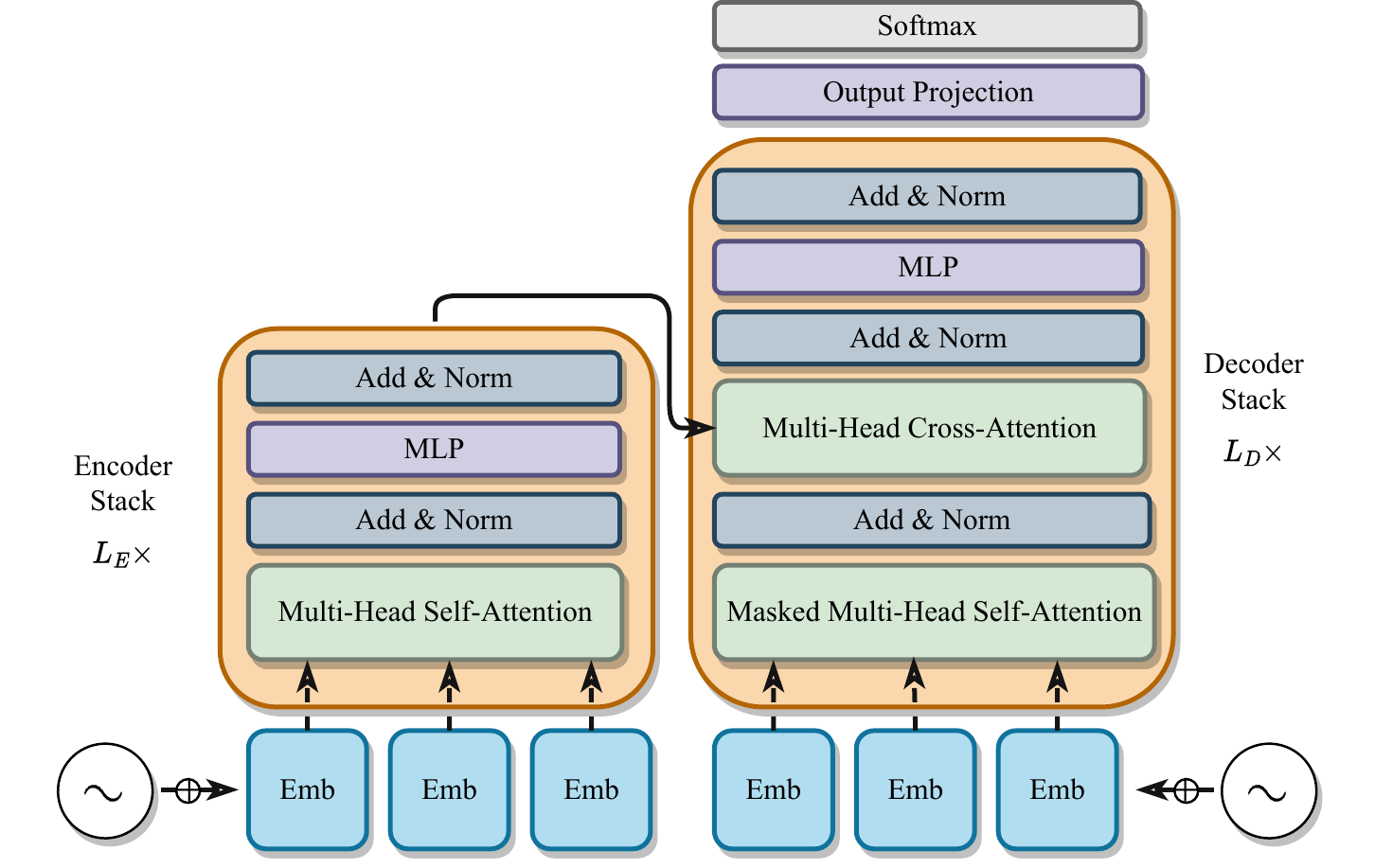}
    \caption{Transformer (\textsc{Tr-Tr})}
    \label{fig:tr_tr}
    \end{subfigure}
    \hfill
    \begin{subfigure}{0.49\textwidth}
    \centering
    \includegraphics[width=\textwidth]{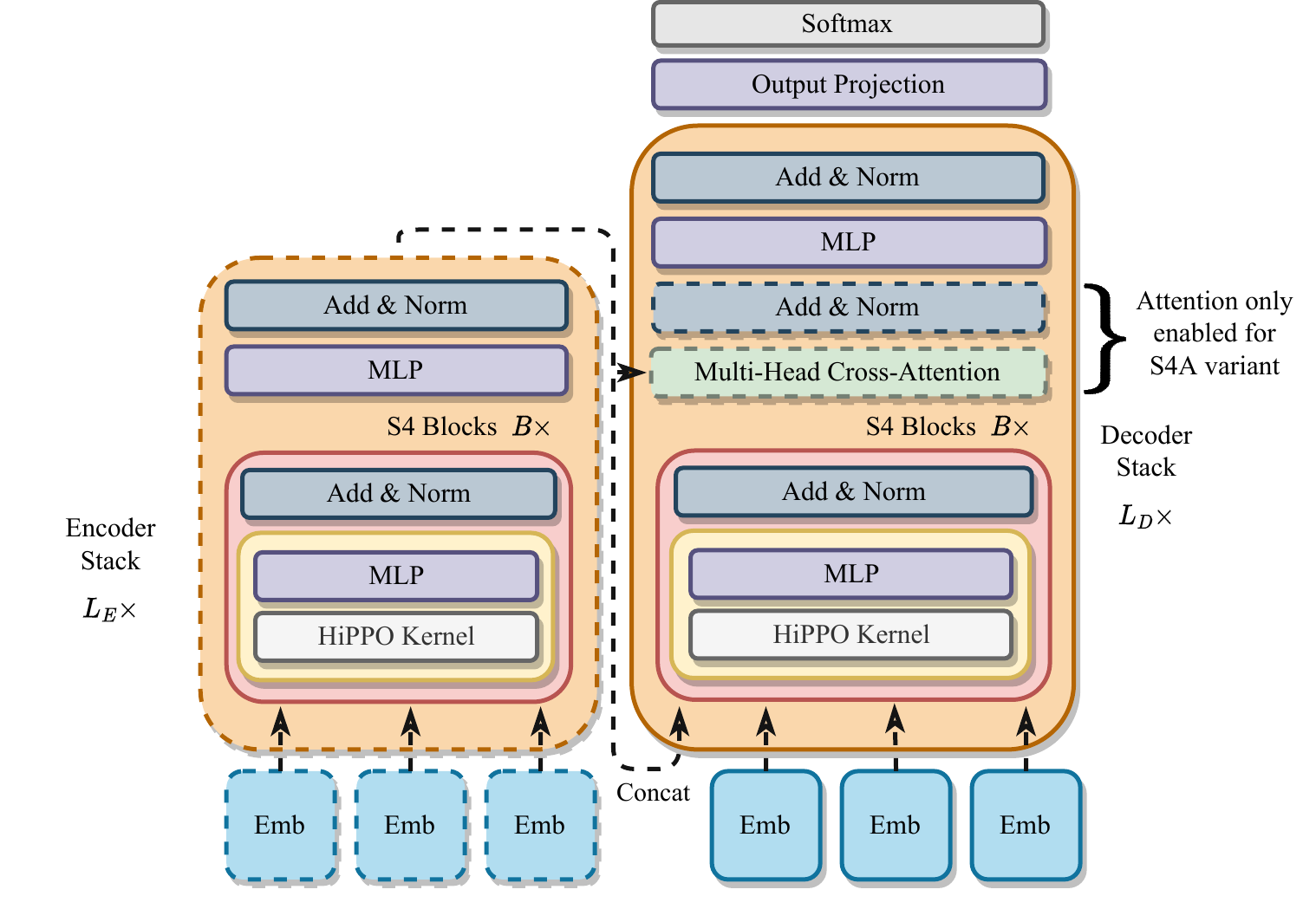}
    \caption{State Spaces (\textsc{S4-S4} and $\emptyset$\textsc{-S4})}
    \label{fig:s4_s4}
    \end{subfigure}
    \caption{Overview of the architectures used. The Transformer architecture (a) is compared to a S4 architecture with an optional encoder (b). ``Add \& Norm'' represents the residual connection and normalization blocks used. The attention module is used only for the \sfoura{} variant (see \Cref{sec:attention_need}).}
    \label{fig:architectures}

\end{figure*}


\section{Background}
\label{sec:background}

In this section, we provide a brief overview of S4 and Machine Translation.

\subsection{Structured State Space Models}
\label{sec:state_spaces}
The continuous state space model (SSM) is defined by:
\begin{equation}
\begin{aligned}
x^{\prime}(t) &=\boldsymbol{A} x(t)+\boldsymbol{B} u(t) \\
y(t) &=\boldsymbol{C} x(t)+\boldsymbol{D} u(t),
\label{eq:state}
\end{aligned}
\end{equation}
where $u(t)$ is a 1D input signal that is mapped to the latent state $x(t)$ and finally to the output $y(t)$. $\boldsymbol{A}$, $\boldsymbol{B}$, $\boldsymbol{C}$, and $\boldsymbol{D}$ are learned parameters. Similar to \newcite{gu2022efficiently}, we assume $\boldsymbol{D}=0$ since it is equivalent to a residual connection.

\paragraph{Discretization}
Following \newcite{gu2022efficiently}, we discretize \Cref{eq:state} to apply it to discrete sequences:
\begin{equation}
\label{eq:discrete_ssm}
\begin{aligned}
x_k &=\overline{\boldsymbol{A}} x_{k-1}+\overline{\boldsymbol{B}} u_k \\
y_k &=\overline{\boldsymbol{C}} x_k,
\end{aligned}
\end{equation}
where $\overline{\boldsymbol{A}} \in \mathbb{R}^{N\times N}$, $\overline{\boldsymbol{B}} \in \mathbb{R}^{N\times 1}$, $\overline{\boldsymbol{C}} \in \mathbb{R}^{1\times N}$ are computed using a bilinear approximation with step size $\Delta$\footnote{Since for Machine Translation the step size does not change, we use $\Delta = 1$.}:
\begin{equation}
\begin{aligned}
\overline{\boldsymbol{A}} &=(\boldsymbol{I}-\Delta / 2 \cdot \boldsymbol{A})^{-1}(\boldsymbol{I}+\Delta / 2 \cdot \boldsymbol{A}) \\
\overline{\boldsymbol{B}} &=(\boldsymbol{I}-\Delta / 2 \cdot \boldsymbol{A})^{-1} \Delta \boldsymbol{B} \\
\overline{\boldsymbol{C}} &= \boldsymbol{C},
\end{aligned}
\end{equation}
and $u(t)$ is sampled at $u_k = u(k\Delta)$.



\Cref{eq:discrete_ssm} is designed to handle 1D input signals. In practice, inputs are rarely 1D, but rather high-dimensional feature vectors, such as embeddings. To handle multiple features, \newcite{gu2022efficiently} use one independent SSM per dimension. These independent SSMs are then concatenated and mixed using a linear layer. For example, if a model has a state size of $64$ and a hidden size of $512$, it will contain $512$ independent SSMs (\Cref{eq:state}). Each of these SSMs has a size of $64$ and processes a single feature. The 1D outputs of these $512$ models are concatenated, and a linear transformation is applied. This process is referred to as an \emph{S4 block}, which involves concatenating all the independent SSMs (one \Cref{eq:discrete_ssm} for each feature), followed by a mixing layer, a residual connection, and Layer Normalization \cite{layernorm:16}.

\paragraph{HiPPO Matrix}
A careful initialization of the $\boldsymbol{A}$ matrix is necessary to reduce exploding/vanishing gradient \cite{gu2022efficiently}. \newcite{GuDERR20} proposed HiPPO-LegS matrices, which allow the state $x(t)$ to memorize the history of the input $u(t)$:
\begin{equation*}
A_{n k} =- \begin{cases}(2 n+1)^{1 / 2}(2 k+1)^{1 / 2} & \text { if } n>k \\ n+1 & \text { if } n=k \\ 0 & \text { if } n<k\end{cases}
\end{equation*}
where $A_{n k}$ is the entry on row $n$ and column $k$. Following \newcite{gu2022efficiently}, we initalize $\boldsymbol{A}$ with the above equation but train it freely afterwards.

\paragraph{Structured State Spaces (S4)}
Finally, \newcite{gu2022efficiently} introduced a set of techniques to make the training of the above architecture more efficient. These include directly computing the output sequence at training time using a single convolution (denoted with $*$):
\begin{equation}
y=\overline{\boldsymbol{K}} * u_k.
\end{equation}
where $\overline{\boldsymbol{K}}$ is a kernel given by:
\begin{equation}
\begin{aligned}
\overline{\boldsymbol{K}} &:=\left(\overline{\boldsymbol{C A}}^i \overline{\boldsymbol{B}}\right)_{i \in[L]} \\
&=\left(\overline{\boldsymbol{C B}}, \overline{\boldsymbol{C A B}}, \ldots, \overline{\boldsymbol{C A}}^{L-1} \overline{\boldsymbol{B}}\right),
\end{aligned}
\end{equation}
and $L$ is the sequence length.
At inference time, \Cref{eq:discrete_ssm} is applied step-by-step. For more details, see \newcite{gu2022efficiently}.

\subsection{Machine Translation (MT)}
\label{seq:s4_mt}

Let $(x_{1:n}, y_{1:m})$ be a source and target sentence pair. The negative log-likelihood of $y$ given $x$ can be written as:
\begin{equation}
    \label{eq:MT}
    \medmath{-\log p(y_{1:m} \mid x_{1:n}) = -\sum_{i=1}^{m} \log p(y_i \mid x_{1:n}, y_{<i})},
\end{equation}
where $p(y_i \mid x_{1:n}, y_{<i})$ is modeled using a neural network.
In encoder-decoder models, such as the Transformer \cite{VaswaniSPUJGKP17}, the model has two main components: an encoder, responsible for capturing source-side dependencies, and a decoder, which captures both target-side and source-target dependencies.

Alternatively, MT can be treated as a Language Modeling task, where the (decoder-only) model is trained on the concatenated source and target sentences, separated with a special \texttt{[SEP]} token in between \cite{wang2021language,gao2022encoder}. Following this approach, the negative log-likelihood is written as:
\begin{align}
    \label{eq:LM4MT}
    \medmath{-\log p(y_{1:m}, x_{1:n}) =}
    & \medmath{\overbrace{-\sum_{j=1}^{n} \log p(x_j \mid x_{<j})}^{\mathcal{L}^{AE}} \quad + \nonumber} \\
    & \medmath{\underbrace{-\sum_{i=1}^{m} \log p(y_i \mid x_{1:n}, y_{<i})}_{\mathcal{L}^{MT}}}.
\end{align}
The $\mathcal{L}^{AE}$ term corresponds to the source reconstruction loss, while $\mathcal{L}^{MT}$ is identical to \Cref{eq:MT}. Since our focus is on MT, we only need to optimize the second term, \ie, $\mathcal{L}^{MT}$. In our experiments, including both loss terms degraded translation quality (see \Cref{app:lm_loss}). Therefore, for our decoder-only models using only the second term, $\mathcal{L}^{MT}$.

\subsection{Transformer}
Transformers \cite{VaswaniSPUJGKP17} are the state-of-the-art architecture for MT. We show a typical architecture in \Cref{fig:tr_tr}.
In particular, both encoder and decoder layers have self-attention and multi-layer perceptron (MLP) modules, and the decoder layer has an extra cross-attention module.

To simplify the text, we will refer to the architectures we discuss as \textsc{[Enc]-[Dec]}, where \textsc{[Enc]} and \textsc{[Dec]} refer to the architecture used. For example, the Transformer model in \Cref{fig:tr_tr} will be referred to as \textsc{Tr-Tr}, since both the encoder and decoder are from the Transformer.

\section{S4 for Machine Translation}


\subsection{Base Architecture}
Following \newcite{gu2022efficiently}, our architectures are based on the Transformer, but with the S4 block (\Cref{sec:background}) replacing self-attention.
In our initial experiments, we intentionally omitted the use of cross-attention in our models to determine whether S4's internal states alone suffice in capturing long-range dependencies for MT.
We call the $B$ consecutive S4 blocks together with the MLP layer, followed by a residual connection and normalization, one \emph{S4 layer}. \newcite{gu2022efficiently} use $B=2$.

We consider two approaches (\Cref{fig:s4_s4}): a decoder-only model ($\emptyset-$\sfour), and an encoder-decoder architecture (\sfour-\sfour).
Our decoder-only model is based on \newcite{gu2022efficiently}, which was shown to perform well in language modeling. This model is designed to predict the next target token by taking as input the concatenated source and the previously predicted target tokens.
Our \sfour-\sfour{} encoder-decoder architecture consists of $L_E$ \sfour{} encoder layers and $L_D$ S4 decoder layers, \emph{without} cross-attention. Instead, we use a simple method to propagate information between the encoder and the decoder: concatenating the encoder outputs with the shifted target sequence. This way, the decoder processes both the encoder outputs and the target tokens.\footnote{Ideally, we would initialize the \sfour{} decoder state spaces with the last state of the encoder. However, this is non-trivial to implement, since the forward step is executed as a single convolution during training. We leave the exploration of this method to future work.}

Finally, for some of the latter experiments, we consider the case where encoder is bidirectional, which we will refer to as \sfourbi{}. In this configuration, the S4 blocks have two sets of parameters ($\overline{\boldsymbol{A}}$, $\overline{\boldsymbol{B}}$ and $\overline{\boldsymbol{C}}$), one per direction.

\subsection{S4 with Cross-Attention}
\label{sec:s4attention}
In our later experiments, we employ a modified S4 decoder architecture, S4A (S4 with Attention). S4A can be used with either a Transformer or S4 encoder. It incorporates a multihead cross-attention module on top of the HiPPO kernel, as shown in \Cref{fig:s4_s4}. Specifically, cross-attention is inserted above the ``Add \& Norm'' layer in the S4 block, followed by another ``Add \& Norm'' layer, similar to the Transformer architecture. When cross-attention is employed, we no longer concatenate the encoder outputs to the shifted target sequence.

\section{Results}
In this section, we describe the experimental setup, and discuss our results.

\subsection{Experimental Setup}
\label{sec:setup}

\paragraph{Data}
We run experiments on WMT'14 English$\leftrightarrow$German (\len$\leftrightarrow$\lde, $4.5$M sentence pairs), and WMT'16 English$\leftrightarrow$Romanian (\len$\leftrightarrow$\lro, $610$K sentence pairs), allowing us to measure performance on four translation directions. For our analysis, we focus on \len$\rightarrow$\lde{}. We tokenize all data using the Moses tokenizer and apply the Moses scripts \cite{koehn-etal-2007-moses} for punctuation normalization. We use Byte-pair encoding (BPE, \newcite{sennrich-etal-2016-neural}) with $40,000$ merge operations, and the WMT'16 provided scripts to normalize \len$\leftrightarrow$\lro{} for the \lro{} side, and to remove diacritics when translating \lro$\rightarrow$\len. Translations into Romanian keep diacritics to generate accurate translations. We evaluate using \sacrebleu\footnote{\url{https://github.com/mjpost/sacrebleu}} version \texttt{2.1.0} \cite{post-2018-call}, with signature \texttt{nrefs:1} \texttt{|} \texttt{case:mixed} \texttt{|} \texttt{eff:no} \texttt{|} \texttt{tok:13a} \texttt{|} \texttt{smooth:exp}. We run all experiments using \textsc{fairseq} \cite{ott-etal-2019-fairseq}, onto which we ported the code from \newcite{gu2022efficiently}\footnote{\url{https://github.com/HazyResearch/state-spaces}}.

Unless stated otherwise, we report \bleu scores on the WMT'14 \len$\rightarrow$\lde{} validation set.

\paragraph{Hyperparameters}
We optimize using \adam \cite{KingmaB14}. After careful tuning, we found the best results with a learning rate of $0.005$ for the S4 models, $0.001$ for the Transformer models, and $0.002$ for the hybrid models. We train for $100$ epochs ($28\,000$ steps), by which point our models had converged, and average the last $10$ checkpoints. We use $4\,000$ warm-up steps and an inverse square root learning rate scheduler \cite{VaswaniSPUJGKP17}. We used a dropout rate of $0.1$ for \len$\leftrightarrow$\lde, and $0.3$ for \len$\leftrightarrow$\lro.
Unless stated otherwise, all models have layer and embedding sizes of $512$, the hidden size of the feed-forward layers is $2048$, and we use $8$ attention heads for the Transformer.
For both the Transformer and S4, we use post-normalization\footnote{In our experiments, we didn't observe any difference between pre and post-normalization.}.
Following \newcite{gu2022efficiently} we use GeLU activation \cite{hendrycks:gelu} after the S4 modules and GLU activation \cite{DBLP:conf/icml/DauphinFAG17} after the linear layer.

\paragraph{S4-specific Training Details}
During our exploration, we experimented with several choices that had a marginal effect on performance:
\begin{enumerate}[label=(\roman*)]
    \item \emph{Module-specific learning rates.} \newcite{gu2022efficiently} suggested different learning rates for the matrices in \cref{eq:discrete_ssm} and the neural layer, but we did not observe any significant difference.
    \item \emph{Trainable $\overline{\boldsymbol{A}}$ and $\overline{\boldsymbol{B}}$.} In line with \newcite{gu2022efficiently}, freezing $\boldsymbol{A}$ and $\boldsymbol{B}$ did not cause a noticeable performance drop.
    \item \emph{State dimension.} We varied the size of the state ($x_k$ in \Cref{eq:discrete_ssm}), but found that that increasing it dimension beyond $64$ did not noticeably affect translation quality. Therefore, similarly to \newcite{gu2022efficiently}, we set the state dimension to $64$ in our experiments. Note that this parameter should not be confused with the model's hidden size, which we examine in \Cref{sec:scaling}. Increasing the state dimension increases the modeling capacity of the S4 kernel for \textbf{each input} dimension, but the output is still collapsed to the hidden size, making the latter the bottleneck.
    \item \emph{Learning rate scheduler.} We observed no significant difference between using the inverse square root scheduler and the cosine scheduler suggested in \cite{gu2022efficiently}.
\end{enumerate}

\subsection{Parameter Allocation and Scaling}
\label{sec:scaling}
\begin{figure*}[t]
    \centering
    \begin{subfigure}{0.31\textwidth}
    \centering
    \includegraphics[width=\textwidth]{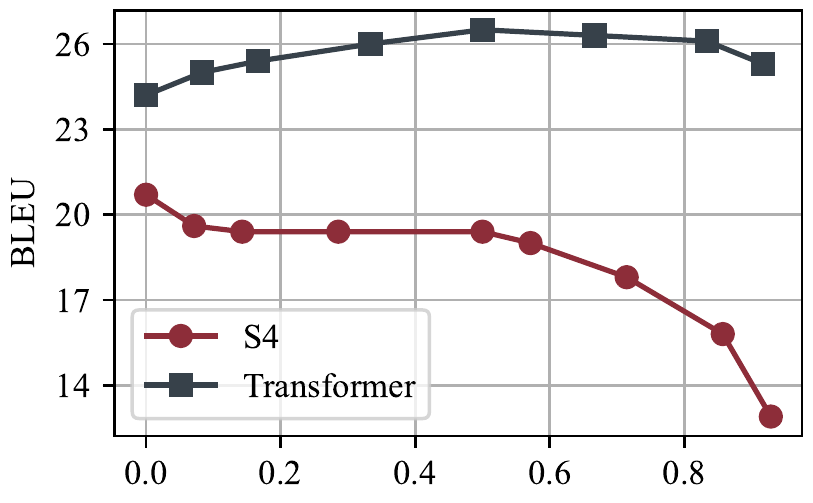}
    \caption{Encoder parameter allocation (\emph{ratio}).}
    \label{fig:scale_ratio}
    \end{subfigure}
    \hfill
    \begin{subfigure}{0.31\textwidth}
    \centering
    \includegraphics[width=\textwidth]{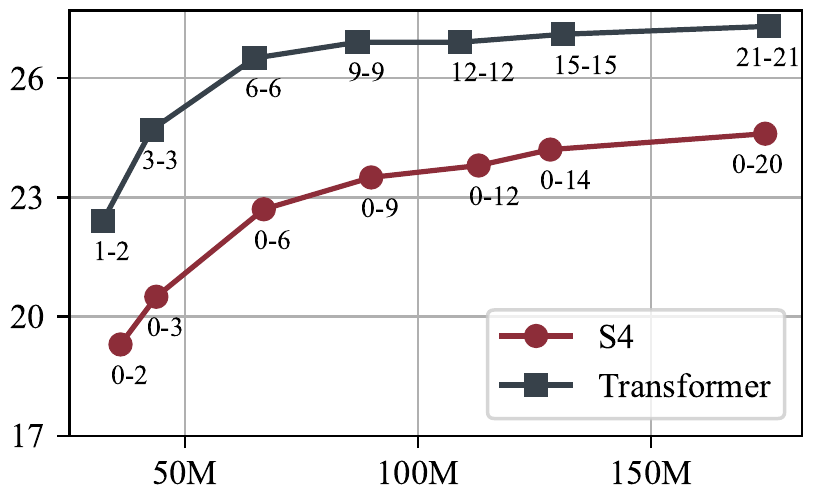}
    \caption{Number of parameters (\emph{depth}).}
    \label{fig:scale_depth}
    \end{subfigure}
    \hfill
    \begin{subfigure}{0.31\textwidth}
    \centering
    \includegraphics[width=\textwidth]{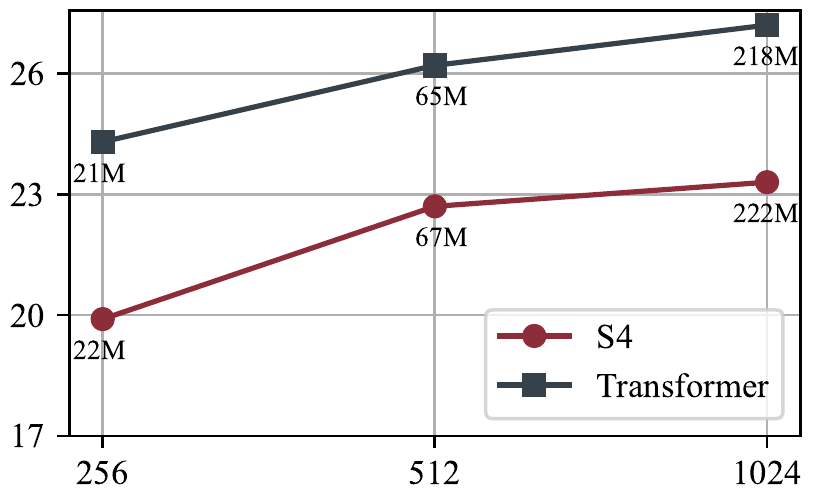}
    \caption{Hidden size (\emph{width}).}
    \label{fig:scale_width}
    \end{subfigure}
    \caption{Scaling plots for S4 and the Transformer. We explore shifting the parameter allocation between the encoder (a), depth scaling (with a fixed hidden size of $512$), symmetrically for the encoder-decoder Transformer, and on the decoder for S4 (b), and hidden size (width) scaling (c), with $0$-$6$ and $6$-$6$ layers of S4 and Transformer, respectively.}
    \label{fig:params}
\end{figure*}

\paragraph{Encoder Scaling}
To explore the effect of parameter allocation on performance, we compare the translation quality of different encoder-decoder configurations with the same total number of parameters (roughly $65$M). In \Cref{fig:scale_ratio}, the $x$ axis represents the ratio of encoder layers to the total number of layers (encoder + decoder). Starting with a decoder-only model ($\text{ratio}=0$), we gradually increase the number of encoder layers, and end with a model containing only a single decoder layer. Two results stand out: first, there is a wide gap between the best S4 and Transformer models: $20.7$ and $26.4$ \bleu, respectively. Second, and consistent with prior work, we find that an even split of parameters between the encoder and decoder ($6$ encoder layers and $6$ decoder layers, \ie, Transformer base) yields the best translation quality for the Transformer \cite{VaswaniSPUJGKP17}, whereas no encoder produces the best results for S4. Based on this finding, we focus on the S4 decoder-only variant for the next experiments.

\paragraph{Number of S4 Blocks per Layer}
Prior research set the number of S4 blocks, $B$, to $2$  \cite{gu2022efficiently}. We found that increasing $B$ is beneficial as S4 blocks are responsible for capturing dependencies between tokens.
In \Cref{tab:s4perlayer} we vary $B$ while keeping the parameter count roughly constant. Increasing $B$ leads to noticeable quality improvements until $B=10$. This architecture achieves a score of $22.7$ \bleu, but the gap to the Transformer is still substantial: $3.7$ \bleu points.
From here onward we use $B=10$ and $6$ layers for the decoder-only model, unless stated otherwise.

\paragraph{Depth Scaling}
In \Cref{fig:scale_depth} we show \bleu{} as we increase the number of layers.
The $x$ axis shows the total number of parameters of each architecture, and the numbers next to each data point indicate the architecture (\eg, $1$-$2$ means a $1$ layer encoder and $2$ layer decoder). There is a clear gap in performance between the two models, which is decreasing as more layers are added, \ie S4 seems to benefit more from increasing the number of layers.

\paragraph{Width Scaling}
In \Cref{fig:scale_width} we examine the influence of the hidden size on both S4 and Transformer, for the $0$-$6$ and $6$-$6$ architectures, respectively. While S4's performance improves with increasing width, the returns are diminishing, and the gap to the Transformer does not go way.

\begin{table}[t]
    \centering
    \small
    \begin{tabular}{llccc}
    \toprule
    $B$ & $L_D$ & ${|\boldsymbol{\theta}_{\text{S4}}|}$ & $|\boldsymbol{\theta}|$ & \textsc{\bleu} \\
    \midrule
     $1$  & $17$ & $10$M & $66$M & $20.0$ \\
     $2$  & $14$ & $20$M & $66$M & $20.7$ \\
     $3$  & $12$ & $21$M & $66$M & $21.2$ \\
     $4$  & $10$ & $23$M & $64$M & $21.5$ \\
     $6$  & $8$  & $28$M & $64$M & $22.1$ \\
     $10$ & $6$  & $35$M & $67$M & $\mathbf{22.7}$ \\
     $16$ & $4$  & $37$M & $65$M & $22.0$ \\
     $22$ & $3$  & $38$M & $64$M & $22.2$ \\
     $35$ & $2$  & $40$M & $64$M & $22.5$ \\
     \bottomrule
    \end{tabular}
    \caption{Effect of number of S4 blocks per layer on the decoder-only architecture. $B$ is the number of S4 blocks, $L_D$ the number of decoder layers, $|\boldsymbol{\theta}_{\text{S4}}|$ are the parameters allocated for S4 inside the HiPPO kernels, and $|\boldsymbol{\theta}|$ are the total parameters.}
    \label{tab:s4perlayer}
\end{table}

\subsection{Translation Quality Comparison}
\label{sec:quality}
Despite our extensive tuning of the S4 architecture, a gap of almost $4$ \bleu points to the Transformer remains.
In this section, we delve deeper into S4's results to determine why it is struggling.


\begin{table}[t]
    \centering
    \small
\begin{tabular}{lcccc}
\toprule
& {Short} & {Medium} & {Long} & {\multirow{2}{*}{Overall}} \\
& {$[1,17]$} & {$[18,29]$} & {$[30,117]$} & \\
\midrule
\tr-\tr      &  $25.9$ &   $26.8$ &    $26.4$ &   $26.4$ \\
\cdashlinelr{1-5}
\sfour-Normal  &  $24.0$ &   $24.3$ &    $21.4$ &   $22.7$ \\
\sfour-Reverse &  $23.2$ &   $24.2$ &    $22.5$ &   $23.1$ \\
\bottomrule
\end{tabular}
    \caption{Translation quality of S4, trained on regular and reversed source sentences, compared to Transformer on the WMT'14 \lende{} validation set, for different reference sentence lengths. Each bucket has approximately $1$k sentences.}
    \label{tab:sentence_length}
\end{table}

\paragraph{Sentence Length}
In \Cref{tab:sentence_length}, we split the source sentences into $3$ buckets according to their length\footnote{To limit spuriousness issues, we chose the buckets so that each bucket has roughly $1$k sentences.}, and show the \bleu scores for both S4 and the Transformer. There is a clear gap between the two models, which increases with sentence length. Specifically, the gap is $1.9$ and $2.5$ \bleu for short and medium-length sentences, respectively, but it increases to $5$ for the longest bucket. This observation is not entirely surprising: S4 uses a fixed-size vector to compress the full source sentence and the previous target tokens, which is not enough for long sentences. The Transformer, on the other hand, has no such constraint, as its attention mechanism lets it retrieve previous states as needed.

\paragraph{Reversing Source Sentences}
To further investigate whether the limited representation size is causing the poor performance of the model, we applied a technique from the earlier neural MT literature.
Before the introduction of attention \cite{DBLP:journals/corr/BahdanauCB14}, it was observed that reversing the source sequence could improve performance by decreasing the distance between cross-language dependencies \cite{SutskeverVL14}.
We trained a model on reversed source sentences, and report the results in \Cref{tab:sentence_length} as S4-Reverse. Compared with the regular model, we get a small overall improvement of $0.4$ \bleu points, but a large improvement of $1.1$ \bleu on long sentences.
This observation suggests that although the HiPPO matrix has promising temporal characteristics, S4 is not able to adequately represent the source sentence and utilize its content during the decoding phase.

\subsection{The Importance of Attention}
\label{sec:attention_need}
In the previous section, we showed that S4 struggles to translate long sentences.
In this section, we study the influence of each source token on the output of the model.

\begin{figure*}[t]
    \centering
    \begin{subfigure}{0.47\textwidth}
    \centering
    \includegraphics[scale=0.6]{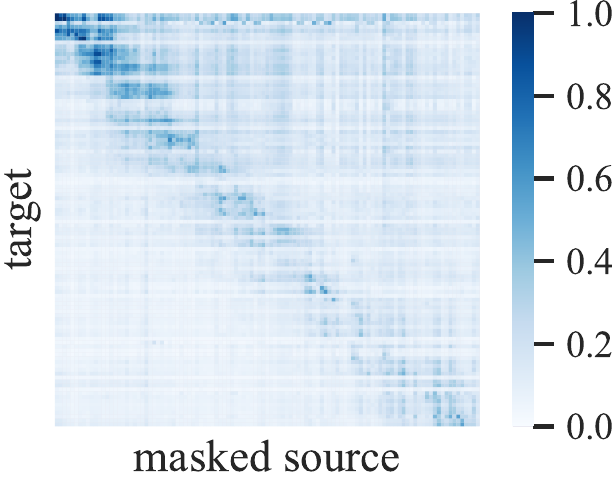}
    \includegraphics[scale=0.6]{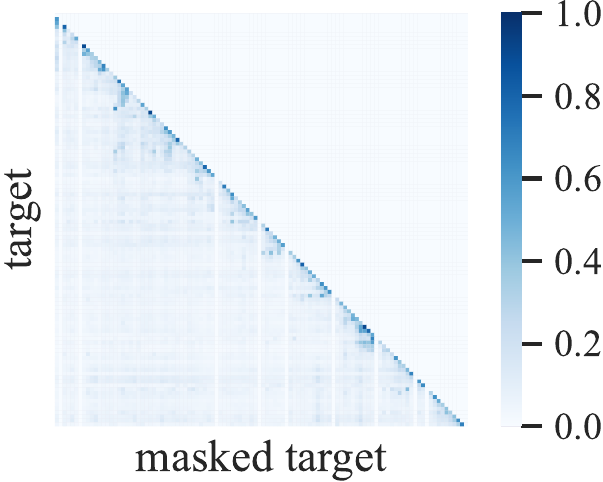}
    \caption{$\emptyset$\textsc{-S4}}
    \label{fig:long_ende_s4s4}
    \end{subfigure}
    \hfill
    \begin{subfigure}{0.5\textwidth}
    \centering
    \includegraphics[scale=0.6]{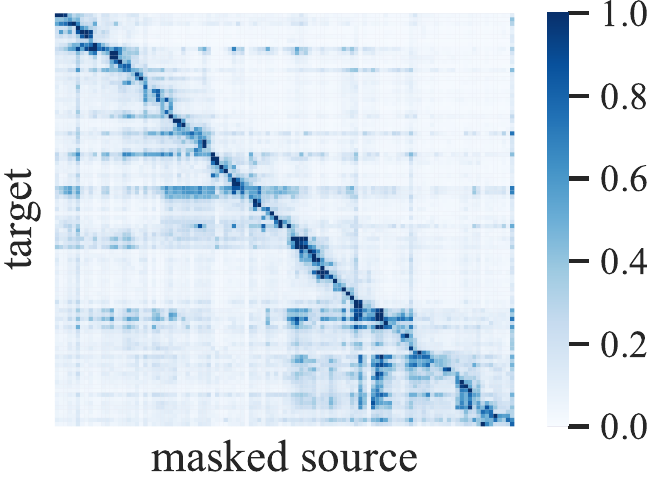}
    \includegraphics[scale=0.6]{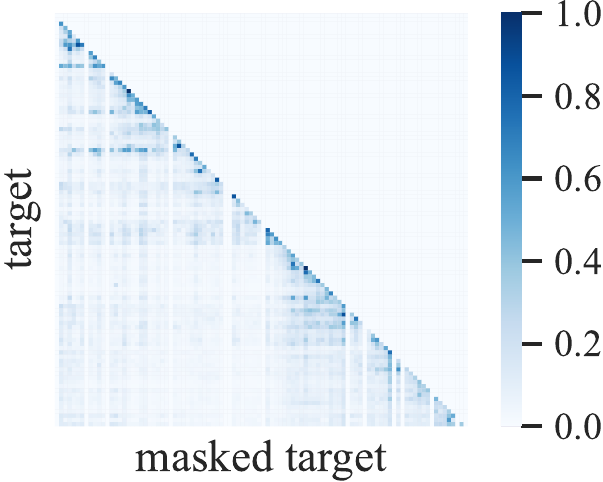}
    \caption{\textsc{\tr-\tr}}
    \label{fig:long_ende_trtr}
    \end{subfigure}
    \caption{Change in the final decoder hidden state for each generated token when masking out source and target tokens in one \emph{long} sample of \lende{} ($109$ tokens), for the decoder-only S4 (a) and the Transformer (b). While the latter can discriminate between source words very accurately (sharp diagonal in b), S4 fails to do so.
    }
    \label{fig:long_ende}
\end{figure*}

\begin{figure*}[t]
    \centering
    \begin{subfigure}{0.49\textwidth}
    \centering
    \vspace{0.5em}
    \includegraphics[scale=0.56]{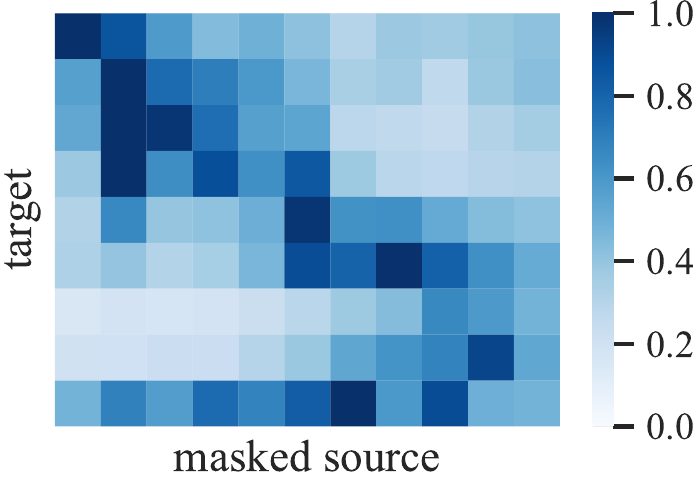}
    \includegraphics[scale=0.56]{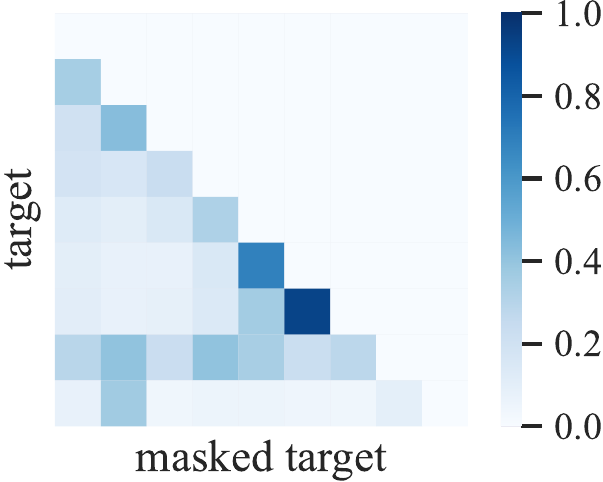}
    \caption{$\emptyset$\textsc{-S4}}
    \label{fig:short_ende_s4s4}
    \end{subfigure}
    \hfill
    \hfill
    \begin{subfigure}{0.49\textwidth}
    \centering
    \vspace{0.5em}
    \includegraphics[scale=0.56]{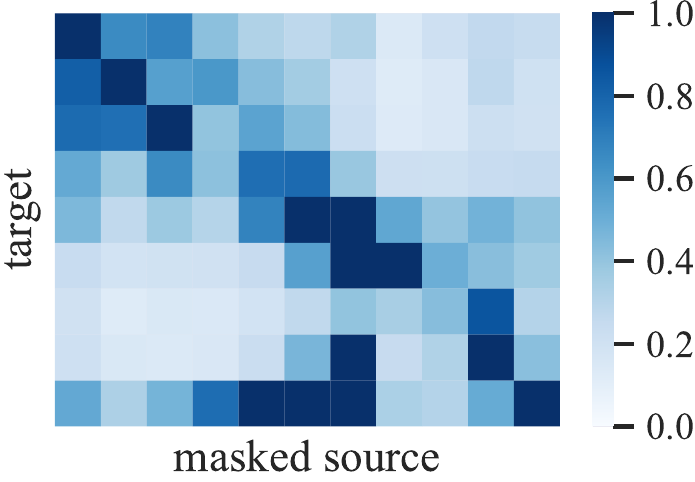}
    \includegraphics[scale=0.56]{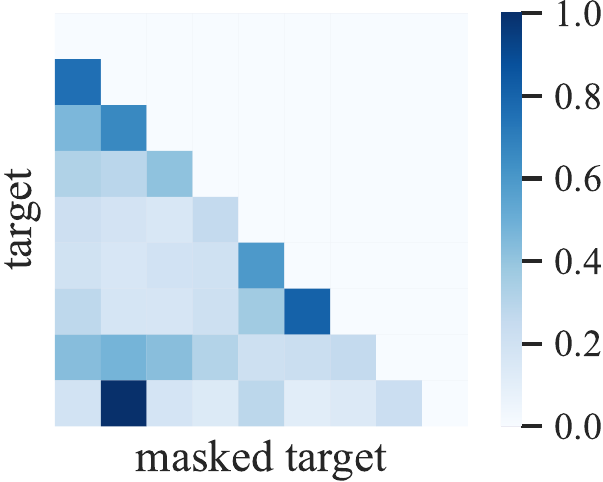}
    \caption{\textsc{TR-TR}}
    \label{fig:short_ende_trtr}
    \end{subfigure}
    \caption{Change in the final decoder hidden state for each generated token when masking out source and target tokens in one \emph{short} sample of \lende{} ($11$ tokens) for the decoder-only S4 and the  Transformer. In the case of short sentences, S4 is able to more accurately align source and target words.}
    \label{fig:short_ende}
\end{figure*}

\paragraph{Attention Heatmaps}
To investigate the extent to which S4 captures dependencies between source and target words, we use a method from \newcite{he-etal-2019-towards}. For each generated target token, we mask out the source tokens, one by one, and replace them with padding tokens. Then, we measure the relative change in the decoder's final layer activation caused by this intervention using L2 distance. By repeating this process for each source token, we obtain a two-dimensional matrix measuring the impact of each source token on each target token. Similarly, we can perform the same procedure by masking the previous target tokens to obtain a similar plot for target-side self-dependencies.

We show the heatmaps for both S4 and the Transformer\footnote{The plots are qualitatively similar to the usual attention weights heatmaps for the Transformer. We show these ``masking'' maps for both models for fair comparison.} in \Cref{fig:long_ende}.
As shown, the differences are stark. The Transformer is focused on just a few words (sharp diagonal in \cref{fig:long_ende_trtr}), while S4 is is much more ``blurred'' and unable to appropriately attend to specific parts of the source sentence.
The difference is not as pronounced for short sentences (see \Cref{fig:short_ende}), indicating that a single hidden state is not enough to capture all the information the model needs for longer sentences.

In \Cref{app:b_heatmaps}, we explore how $B$ impacts the heatmaps. We find that increasing $B$ sharpens the heatmaps, although they never get as sharp as those of the Transformer.

\subsection{Attention-enhanced Architectures}
In the previous experiments, we found that S4 underperforms on long sentences, and hypothesized that this is due to its fixed-size representation, which makes it unable to recall the full source sentence. To address this, we now extend the S4 decoder with an attention mechanism, which allows us to use an encoder-decoder setup, \sfour-\sfoura. For more details on the attention mechanism,  see \Cref{sec:s4attention}.

We conducted experiments similar to those in \Cref{sec:scaling} to determine the optimal $B$ and how to allocate layers to the encoder and the decoder, while keeping the total number of parameters constant. We summarize the findings in \Cref{tab:s4a_layers,tab:s4a_encdec_allocation}. We found the best results with a balanced architecture, $5-5$, and $B=3$.
This model improves performance by almost $3$ \bleu points on the WMT'14 validation set, from $22.7$ to $25.6$.
From here onward, encoders and decoders have $5$ layers for S4 and $6$ layers for Transformer.

\begin{table}[t]
    \centering
    \small
    \begin{tabular}{ccccc}
    \toprule
    $B$ & $L_E$ & $L_D$ & $|\boldsymbol{\theta}|$ & \textsc{\bleu} \\
    \midrule
     $2$  & $6$ & $6$ & $66$M & $24.9$ \\
     $3$  & $5$ & $5$ & $64$M & $\mathbf{25.4}$ \\
     $5$  & $4$ & $4$ & $64$M & $\mathbf{25.4}$ \\
     $8$  & $3$ & $3$ & $63$M & $25.2$ \\
     \bottomrule
    \end{tabular}
    \caption{Effect of number of $B$ and number of encoder ($L_E$) and decoder ($L_D$) layers for the \sfour-\sfoura{} encoder-decoder architecture.}
    \label{tab:s4a_layers}
\end{table}

\begin{table}[t]
    \centering
    \scriptsize
    \setlength{\tabcolsep}{3pt}
    \begin{tabular}{cccccccccc}
    \toprule
    $L_E$ & $1$ & $2$ & $3$ & $4$ & $5$ & $6$ & $7$ & $8$ & $9$ \\
    \cdashlinelr{1-10}
    $L_D$ & $9$ & $8$ & $7$ & $6$ & $5$ & $4$ & $3$ & $2$ & $1$ \\
    \midrule
    \bleu & $24.5$ & $24.8$ & $25.1$ & $25.1$ & $\mathbf{25.4}$ & $25.1$ & $25.1$ & $25.1$ & $23.7$ \\
     \bottomrule
    \end{tabular}
    \caption{Effect of allocating layers to the encoder or to the decoder on the \sfour-\sfoura{} architecture, with $B=3$. The models have a total of $10$ layers between the encoder and decoder.}
    \label{tab:s4a_encdec_allocation}
\end{table}

\begin{table}[t]
    \centering
    \small
\begin{tabular}{lcccc}
\toprule
& {Short} & {Medium} & {Long} & {\multirow{2}{*}{Overall}} \\
& {$[1,17]$} & {$[18,29]$} & {$[30,117]$} & \\
\midrule
$\emptyset$-\sfour &  $24.0$ &   $24.3$ &    $21.4$ &   $22.7$ \\
\tr-\tr          &  $\mathbf{25.9}$ &   $26.8$ &    $26.4$ &   $26.4$ \\
\midrule
\sfour-\tr          &  $24.7$ &   $25.5$ &    $25.2$ &   $25.2$ \\
\sfour-\sfoura         &  $25.0$ &   $26.5$ &    $25.3$ &   $25.6$ \\
\midrule
\sfourbi-\tr        &  $25.5$ &   $25.9$ &    $25.6$ &   $25.7$ \\
\sfourbi-\sfoura       &  $25.3$ &   $26.5$ &    $25.8$ &   $25.9$ \\
\midrule
\tr-\sfour          &  $24.2$ &   $24.8$ &    $22.9$ &   $23.7$ \\
\tr-\sfoura         &  ${25.6}$ &   $\mathbf{26.9}$ &    $\mathbf{26.5}$ &   $\mathbf{26.5}$ \\
\bottomrule
\end{tabular}
    \caption{Translation quality of different attention-enhanced models on the WMT'14 \lende{} validation set for different source sentence lengths. Each bucket has approximately $1$k sentences.
    All models have $64M<|\boldsymbol{\theta}|<66M$ parameters.}
    \label{tab:hybrid_sentence_length}
\end{table}

In \Cref{tab:hybrid_sentence_length} we compare the performance of \sfour-\sfoura{} and the Transformer (\tr-\tr) for short, medium, and long sentences. Although there is a noticeable improvement over the attention-free S4 model ($\emptyset$-\sfour), especially for longer sentences, there is still gap between the two models.
One possible explanation for the comparatively poorer performance of \sfour-\sfoura{} is the unidirectional nature of the S4 encoder. This results in subpar representations for the initial words in the source sentence. Indeed, when using a S4 encoder with a Transformer decoder (\sfour-\tr), the performance is still behind that of \tr-\tr, and replacing the S4 encoder with a Transformer (\tr-\sfoura) allows us to match the performance Transformer. Making the S4 encoder bidirectional (\sfourbi), we are able to narrow the performance gap to the Transformer to just $0.5$ \bleu points (see \sfourbi-\sfoura).

\begin{figure}[t]
    \centering
    \centering
    \vspace{0.5em}
    \includegraphics[scale=0.6]{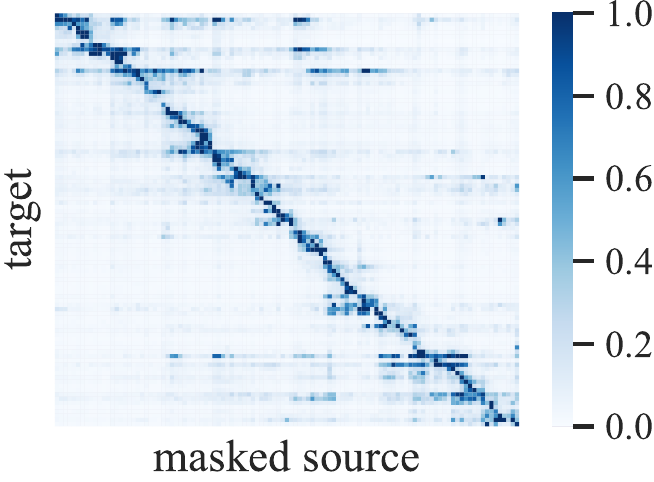}
    \includegraphics[scale=0.6]{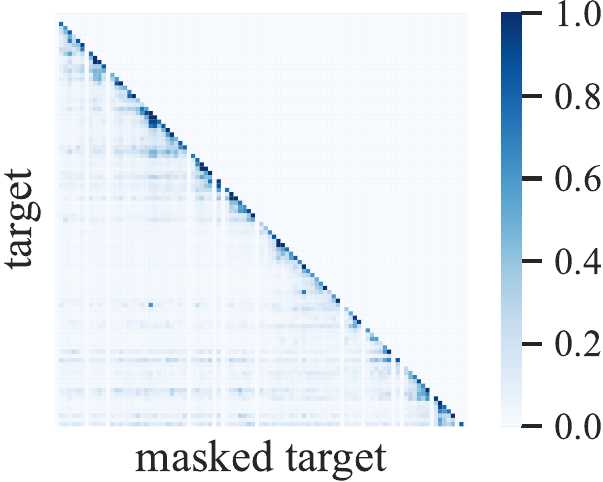}
    \caption{Comparison of \tr-\sfoura's change in the final decoder hidden state for each generated token when masking out source tokens for one \emph{long} sample of \lende ~(the same sample as \Cref{fig:long_ende}). Enhancing S4 with attention helps it to focus on the source tokens, similar to \textsc{Tr-Tr}.}
    \label{fig:long_ende_hybrid}
\end{figure}

Finally, in \Cref{fig:long_ende_hybrid} we show the attention heatmaps for \tr-\sfoura{} architecture, which were generated in the same was as those in \Cref{fig:long_ende}. These plots show that the model is now capable of accurately aligning source and target words, and are qualitatively similar to those of the Transformer.

\paragraph{Why does S4 perform well on LM but not MT?} A natural question to ask is why does S4 perform well on LM \cite{gu2022efficiently}, but not on MT. Our intuition is that MT is a more challenging task. For LM, the model only needs to consider a shorter context to accurately predict the next token, whereas for MT, it requires accurate access to the source sentence representations. As the length of the source sentence increases, a fixed-size state is insufficient to capture fine-grained representations of the source, and thus the model's performance suffers.
This is in line with the observations made by \newcite{vig2019analyzing}, who argue that Transformer LMs tend to pay more attention to the previous few tokens, emphasizing the importance of short-term memory over long-term memory.

\subsection{Results for Other Language Pairs}
In the previous sections, we focused on \lende{}. In this section, we compare the different S4 architectures for other language pairs (\lde-\len, \len-\lro, and \lro-\len) and summarize the results in \Cref{tab:languagepair}. These numbers are on the test sets of the respective language pairs.
The results align with our previous findings. Without attention, there is a significant gap between S4 and the Transformer models, which is reduced significantly by adding it. Interestingly, the best performing architecture for all language pairs is the hybrid \tr-\sfoura{}, which provides a small but statistically significant\footnote{We performed statistical significance tests using paired bootstrap resampling \cite{koehn-2004-statistical} and a significance of $5\%$.} improvement over the Transformer for all but \lde$\rightarrow$\len.

\begin{table}[!tbp]
    \centering
    \small
    \begin{tabular}{lcccc}
    \toprule
    & \lende & \ldeen & \lenro & \lroen \\
    \midrule
    $\emptyset$-\sfour & $22.1\phantom{^\dagger}$ & $25.4$ & $12.8\phantom{^\dagger}$ & $19.7\phantom{^\dagger}$ \\
    \sfourbi-\sfoura & $26.1\phantom{^\dagger}$ & $29.5$ & $22.7\phantom{^\dagger}$ & $31.0\phantom{^\dagger}$ \\
    \tr-\sfoura & $\mathbf{27.3}^\dagger$ & $\mathbf{31.4}$ & $\mathbf{24.1}^\dagger$ & $\mathbf{33.6}^\dagger$ \\
    \tr-\tr & $26.9\phantom{^\dagger}$ & $\mathbf{31.4}$ & $23.8\phantom{^\dagger}$ & $33.2\phantom{^\dagger}$ \\
    \bottomrule
    \end{tabular}
    \caption{\bleu scores on test set for each architecture in $4$ different language pairs. The $\dagger$ on \tr-\sfoura{} indicates statistically significant results.}
    \label{tab:languagepair}
\end{table}

\section{Conclusion and Future Work}
In this work, we explored the application of S4 to Machine Translation and conducted an investigation into the best architecture and hyperparameters. Despite our efforts, we found that S4's translation accuracy lagged behind the Transformer, and the performance gap widened for longer sentences. We then showed that this was due to the limitations of the fixed-size representation used by S4, which had to compress the entire prior context, including the source sentence and previous output tokens. Finally, we showed that the performance gap can be closed by incorporating attention.

Since we did our investigation into S4, numerous new SSM models have been proposed. Of particular note are S5 \cite{smith2023simplified}, which utilizes a multi-input multi-output SSM, instead of one single-input single-output SSM per feature as S4 does, and H3 \cite{dao2023hungry}, which is faster and better at LM than S4. We hope future research explores how well these models perform on MT. Additionally, it is worth noting \textsc{Mega} \cite{ma2023mega}, which incorporates SSM's into the Transformer attention, and is effective in MT, albeit at the expense of quadratic complexity.

\section{Acknowledgements}
We would like to thank António V. Lopes, Hendra Setiawan, and Matthias Sperber for for their suggestions and feedback. Their contributions significantly improved the final work.

\bibliographystyle{eamt23}


\appendix
\newpage
\section{Influence of $\mathcal{L}^{AE}$}
\label{app:lm_loss}

In our experiments with the decoder-only architecture, we intentionally excluded the loss term $\mathcal{L}^{AE}$ from \Cref{eq:MT} as it is not necessary for MT. In \Cref{tab:lm_loss} we show the effect of including this loss during training: performance degradation of around $4$ \bleu points for both architectures.

\begin{table}[h]
    \centering
    \small
    \begin{tabular}{llcccc}
    \toprule
    $B$ & $L_D$ & $|\boldsymbol{\theta}|$ & w/ $\mathcal{L}^{AE}$ & w/o $\mathcal{L}^{AE}$ \\
    \midrule
     $6$ & $8$ & $65$M & $17.9$ & $22.3$ \\
     $10$ & $6$ & $68$M & $18.6$ & $22.5$ \\
     \bottomrule
    \end{tabular}
    \caption{Impact of the autoencoder loss ($\mathcal{L}^{AE}$) on translation quality  on the WMT'14 validation set for two decoder-only architectures. $B$ is the number of S4 blocks, $L_D$ the number of decoder layers (this is a decoder-only architecture), and $|\boldsymbol{\theta}|$ is the number of parameters.}
    \label{tab:lm_loss}
\end{table}

\section{Effect of $B$ in the Cross-Attention Heatmaps}
\label{app:b_heatmaps}

\begin{figure*}[t]
    \centering
    \begin{tabular}{ccc}
    \subfloat[$B=1$ \& $L_D=17$.]{\includegraphics[scale=0.6]{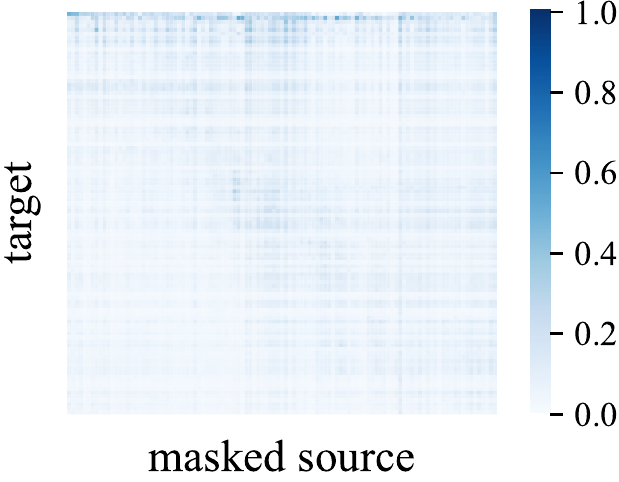}} &
    \subfloat[$B=2$ \& $L_D=14$.]{\includegraphics[scale=0.6]{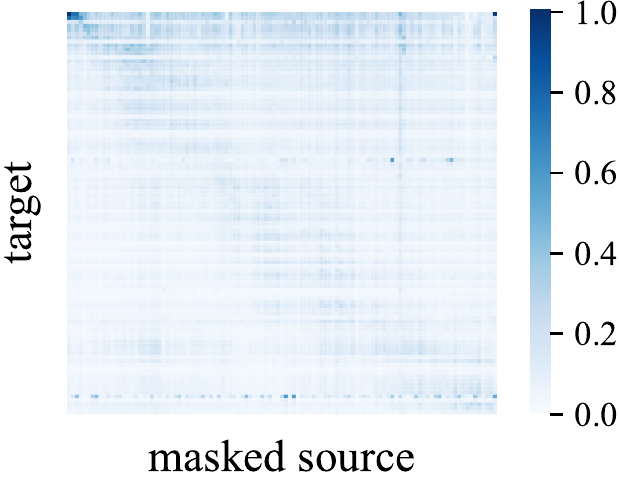}} &
    \subfloat[$B=3$ \& $L_D=12$.]{\includegraphics[scale=0.6]{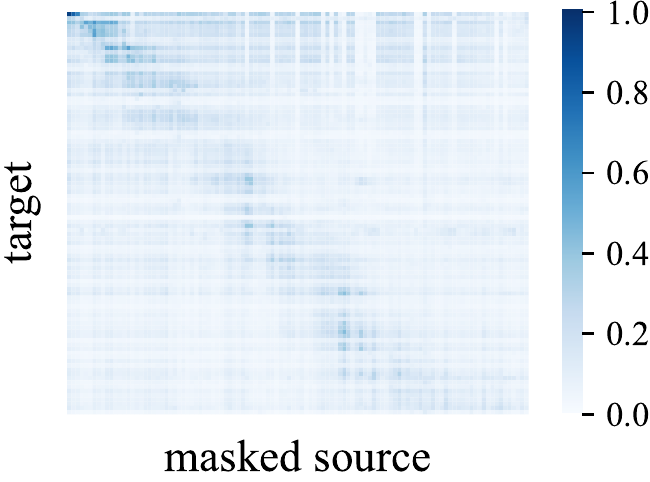}} \\
    \subfloat[$B=4$ \& $L_D=10$.]{\includegraphics[scale=0.6]{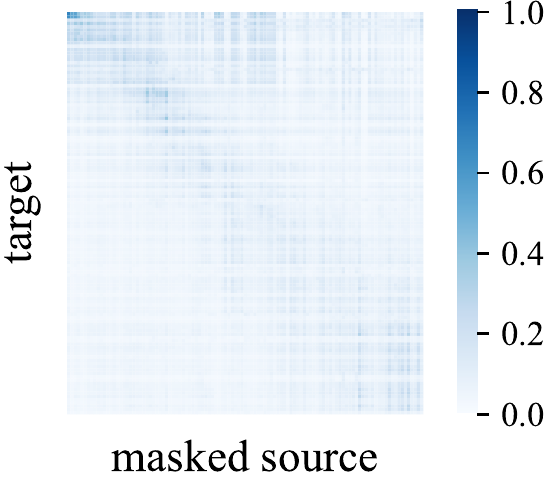}} &
    \subfloat[$B=6$ \& $L_D=8$.]{\includegraphics[scale=0.6]{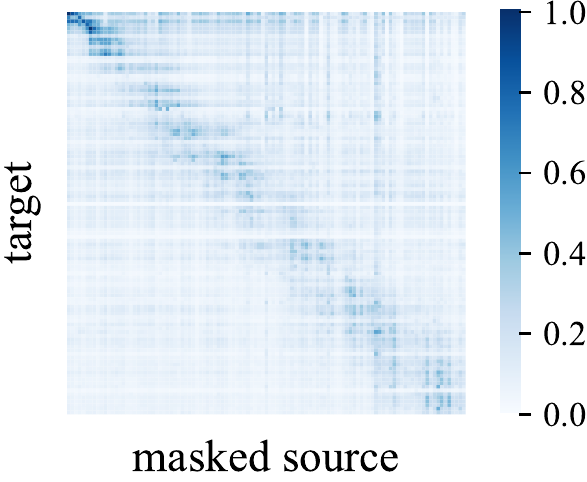}} &
    \subfloat[$B=10$ \& $L_D=6$.]{\includegraphics[scale=0.6]{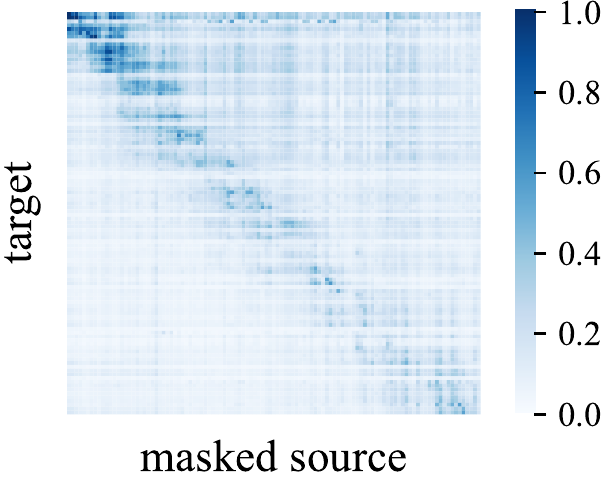}} \\
    \subfloat[$B=16$ \& $L_D=4$.]{\includegraphics[scale=0.6]{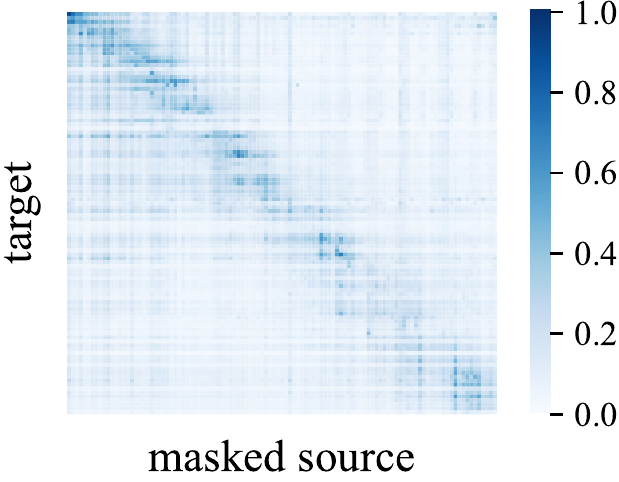}} &
    \subfloat[$B=22$ \& $L_D=3$.]{\includegraphics[scale=0.6]{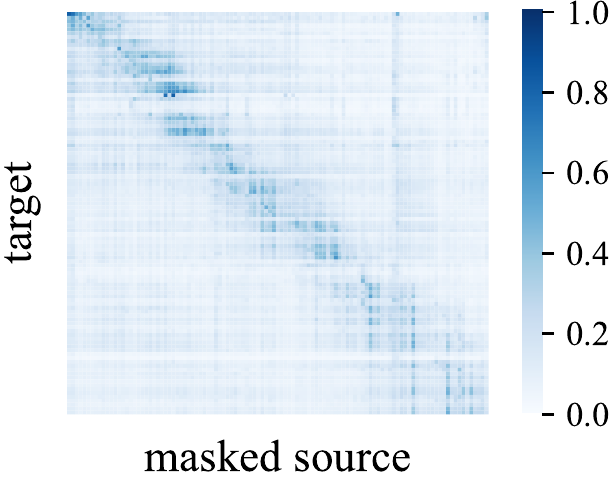}} &
    \subfloat[$B=35$ \& $L_D=2$.]{\includegraphics[scale=0.6]{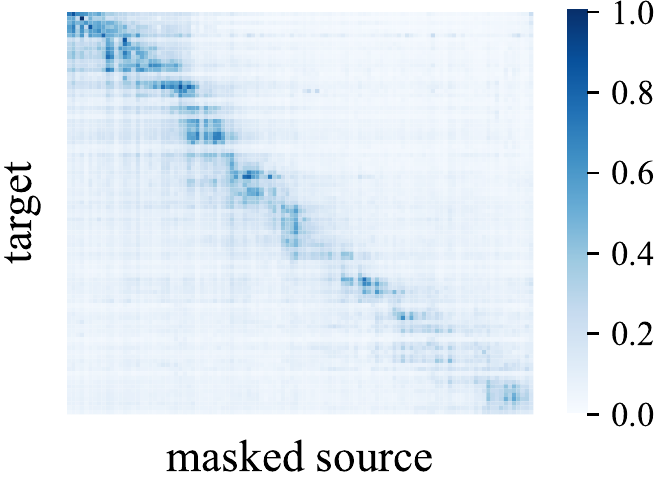}} \\
    \end{tabular}
    \caption{Cross-attention heatmaps for the models in \Cref{tab:s4perlayer}. Increasing $B$ (while keeping the total number of parameters roughly constant) makes the heatmaps less blurry, which means it is easier for the model to retrieve source states.}
    \label{fig:b_ablation_heatmap}
\end{figure*}

Using the methodology described in \Cref{sec:attention_need}, \Cref{fig:b_ablation_heatmap} shows the cross-attention heatmaps for the models in \Cref{tab:s4perlayer}. All models have roughly the same number of parameters, and differ only in $B$ and the number of layers ($L_D$). As in \Cref{fig:long_ende}, the source sentence has $109$ tokens. A noticeable pattern emerges: as $B$ increases, the heatmap sharpens, meaning it is easier for S4 to retrieve the source states. It is worth noting, however, that these heatmaps never get as sharp as those of the models with attention.

\end{document}